\documentclass[11pt]{article}

\usepackage{microtype} 
\usepackage{booktabs}  
\usepackage{url}  
\usepackage{natbib}
 \usepackage{graphicx}
 \usepackage{float}
\usepackage{amsmath}
\usepackage{amsthm}
\usepackage{xcolor}

\colorlet{kwcolor}{violet!80}
\colorlet{akcolor}{blue!80}
\colorlet{jpcolor}{pink}
\colorlet{mwcolor}{red!80}
\colorlet{dkcolor}{green}

 \usepackage[preprint]{automl}
\usepackage{natbib}
\title{Deciphering AutoML Ensembles: \textit{cattleia}'s Assistance in~Decision-Making}

\author[1]{\nameemail{Anna Kozak}{anna.kozak@pw.edu.pl}}
\author[1]{\nameemail{Dominik Kędzierski}{dominikkedzierski0607@gmail.com}}
\author[1]{\nameemail{Jakub Piwko}{jakub.piwko2.stud@pw.edu.pl}}
\author[1]{\nameemail{Malwina Wojewoda}{malwina.wojewoda.stud@pw.edu.pl}}
\author[1]{\nameemail{Katarzyna Woźnica}{katarzyna.woznica@pw.edu.pl}}
\affil[1]{Warsaw University of Technology}

\hypersetup{%
  pdfauthor={}, 
  pdftitle={},
  pdfsubject={},
  pdfkeywords={}
}

\begin{document}

\maketitle

\begin{abstract}
In many applications, model ensembling proves to be better than a single predictive model. Hence, it is the most common post-processing technique in Automated Machine Learning (AutoML). The most popular frameworks use ensembles at the expense of reducing the~interpretability of the final models. In our work, we propose \textit{cattleia} - an application that deciphers the ensembles for regression, multiclass, and binary classification tasks. This tool works with models built by three AutoML packages: auto-sklearn, AutoGluon, and~FLAML. The given ensemble is analyzed from different perspectives. We conduct a predictive performance investigation through evaluation metrics of the ensemble and its component models. We extend the validation perspective by introducing new measures to assess the diversity and complementarity of the model predictions. Moreover, we apply explainable artificial intelligence (XAI) techniques to examine the importance of variables. Summarizing obtained insights, we can investigate and adjust the weights with a modification tool to tune the ensemble in the desired way. The application provides the aforementioned aspects through dedicated interactive visualizations, making it accessible to a diverse audience.
We believe the \textit{cattleia} can support users in decision-making and deepen the comprehension of AutoML frameworks.

\end{abstract}





\section{Introduction} 
In many machine learning applications, the priority is to create accurate but also robust and generalizable models. For such a challenge, ensembles of predictive models prove to be highly competent over single models~\citep{Polikar_Emsemble}.  So, these ensembles are commonly integrated into Automated Machine Learning (AutoML) packages, which strive to generate the most precise models attainable~\citep{Escalante2021}.

The key to increasing the predictive accuracy of the ensembling method is to use a diverse set of models~\citep{Kuncheva_diversity}. This involves favoring models that predict the target differently. Thus, an ensemble composed of those models gains greater prediction flexibility and generalization from multiple predictors~\citep{Rokach_ensemble}. Therefore, the promising scenario would involve different algorithms with different hyperparameters. There are many approaches to~achieve this diversity, from iterative approaches defined in~\citep{caruna1, caruna2} to~pruning methods like~\citep{DAI201775}. Apart from these diverse approaches, there are also basic methodologies for constructing diverse ensembles of models such as boosting~\citep{Freund1996ExperimentsWA_boosting}, bagging~\citep{breiman96_bagging}, and stacking~\citep{Ting_stacking}.

Although ensemble methods in AutoML are powerful tools, the question arises as to whether improved results can be achieved while ensuring the interpretability of models. The challenge lies in comprehending the essence of model diversity and their interdependencies. There is a notable interest in understanding complex models~\citep{Zhou2021EvaluatingTQ} and explainable machine learning is crucial in supporting the decision-making process~\citep{why_trust}, thereby building trust in AutoML models and effectively utilizing them~\citep{trust_in_automl}. Recent studies highlight the potential for elevating prediction accuracy and enhancing the overall credibility of the results thanks to skillfully presented insights~\citep{viz_interpreting}. 
While many applications and visualizations support the examination of the machine learning process, most of them focus on post-modeling tools, with little attention given to ensemble models~\citep{taxonomy}. 

\textbf{Contribution:} In this paper, we present \textit{cattleia} - Complex Accessible Transparent Tool for~Learning Ensembles in AutoML, which aims to bridge these gaps and further advance the field of~AutoML explanations. It is a Dash web application that enhances the interpretability of models built using AutoML packages by offering innovative solutions for ensemble analysis. \textit{cattleia} is compatible with three popular AutoML packages such as \texttt{AutoGluon}~\citep{agtabular}, \texttt{auto-sklearn}~\citep{autosklr}, and \texttt{FLAML}~\citep{flaml}.
 The application provides analysis from four different angles: metrics evaluating individual models and ensemble, compatimetrics examining relationships between models, weights assigned to particular models in the ensemble, and explainable artificial intelligence (XAI) methods assessing the importance of individual variables. 
 The analysis spans from examining the entire ensemble model through pairs of models within it to individual models. It can also be narrowed down to analyzing single variables and specific observations.
 It~supports the data scientist's interaction with already established AutoML frameworks. What is more, \textit{cattleia} provides a set of visualizations and metrics to reduce entry-level in exploring AutoML solutions.

\section{Related Works}

Existing AutoML frameworks present model performance in very different ways, so contrasting them is challenging and needs to be improved. Several tools have been developed for this purpose, mainly focusing on the model creation process in AutoML frameworks. 

The first is \texttt{ATMSeer}~\citep{atmseer}. It supports users in monitoring an ongoing AutoML process, analyzing the construction of the searched models, and refining the search space in real time through a multi-granularity visualization.

The second package for interactive visualization is \texttt{PipelineProfiler}~\citep{pipelineprofiler}, which is integrated with Jupyter Notebook. It focuses on the exploration and comparison of machine learning pipelines generated by various AutoML systems. It accepts pipelines represented with architecture and metadata information, presenting them in a pipeline matrix format that summarises their structure and performance. 

 An interactive visual analytics tool created based on the needs of a diverse potential user group is also \texttt{XAutoML}~\citep{XAutoML}. This solution allows users to compare pipelines, analyze the optimization process, inspect individual models, and evaluate ensembles. It is integrated with JupyterLab as an extension for a user-friendly experience. An interesting feature is a hyperparameter importance visualization, helping users understand the significance of hyperparameters.

Another tool is \texttt{AutoAIViz}~\citep{autoaiviz}. It is a system aiming to visualize the model generation process in AutoML. It provides a real-time overview of the machine learning pipelines as they are being created and the ability to view detailed information at each step in the AutoMl model-building process.

Similar features to those are present within the package \texttt{DeepCAVE}~\citep{deepcave}. It is an interactive framework for analyzing and monitoring the optimization procedures of AutoML. It offers an interactive app for real-time visualization and analysis of AutoML processes. Its modular and extensible design allows for exploration across various domains, including performance analysis, hyperparameter evaluation, and budget assessment. \texttt{DeepCAVE} is invaluable for identifying outliers, comparing multiple executions, and gaining insights into the optimization process.

These studies on explanations of AutoML models have been published recently, illustrating the field's gradual growth. Nonetheless, all the work has focused mainly on studying the model-building process. More tools are needed to allow a comprehensive check of the outcomes of the built models and a comparison of the performance of the models from which the ensembles are formed.

\section{\textit{cattleia}} \label{sec: framework}

In this paper, we introduce \textit{cattleia}, the application written on top of Dash framework for analyzing model ensembles created by popular AutoML packages in Python, namely \texttt{auto-sklearn}, \texttt{AutoGluon} and \texttt{FLAML}. The application and launch instructions are available on the GitHub repository \footnote{\url{https://anon-github.automl.cc/r/cattleia-DC83}} and is provided open source, under the Apache License, Version 2.0.

The visualizations in \textit{cattleia} are generated using the \texttt{Plotly} library~\citep{plotly} due to its ability to generate interactive visualizations and functionalities like zooming and filtering, enabling a~more thorough analysis. \textit{cattleia} does not necessitate the training of models from scratch. Instead, it performs analyses on pre-trained models and makes predictions based on the provided data, ensuring the efficient and smooth application performance. One of the distinguishing features of~this application is customizability since users can incorporate new metrics and packages as per their requirements.

\begin{figure}[H]
    \centering
    \includegraphics[width=\textwidth]{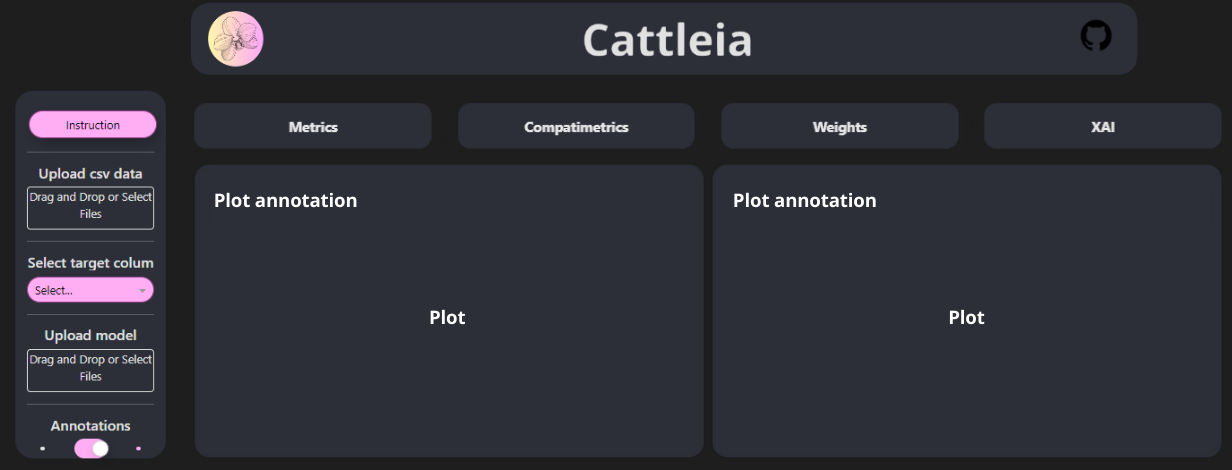}
    \caption{Application interface scheme. There are available four tabs related to different aspects of ensemble analysis. On the left is a sidebar where the user finds instructions and a place to upload a considered ensemble. }
    \label{fig: app schema}
\end{figure}

The functionalities and usage of the application are explained through a user-selectable instructional guide located in the sidebar. To correctly utilize this tool, it is necessary to supply both the data and the model generated using the packages mentioned above and saved in a particular format. The annotations switch permits the display of descriptions, which can be especially useful for deciphering the visualizations. Once the required elements are uploaded, an interactive dashboard is presented, as shown in Figure \ref{fig: app schema}. 

The tabs available represent distinct scopes of ensemble analysis:
\begin{itemize}
    \item \textbf{Metrics} encompass a comparison of evaluation metrics of both component models and ensemble. Depending on whether the model addresses a classification or regression problem, these graphs present the corresponding metrics. In addition, there is a correlation matrix of each model prediction, and the plot compares every single prediction of each model with the actual target value.
    \item \textbf{Compatimetrics} tab is designed to evaluate the models' similarity and their joined performance.
    Compatimetrics are novel measures of model compatibility based on simple heuristics and evaluation metrics; their examples are described in Appendix~\ref{appendix: compatimetrics}. This feature allows for a more in-depth analysis by discovering concealed patterns among models, identifying groups of models that perform well together, and highlighting models that could potentially undermine the predictions. The compatimetrics plots are task-specific, as the metrics used for regression, binary, and multiclass classification differ. 
    \item \textbf{Weights analysis} comprehensively examines how each component model contributes to the overall ensemble score. This tab is designed explicitly for \texttt{AutoGluon} and \texttt{auto-sklearn} packages, as \texttt{FLAML} does not utilize weights when creating ensembles. This functionality uses interactive sliders to adjust the influence of individual models in the ensemble prediction. In the regression task, it is a weighted average of individual model predictions, and for classification, it is a weighted average of predicted probabilities guiding class decisions. This approach enables users to review the metrics of such custom weighted models through an automatically updating table.
    \item The \textbf{XAI} tab allows us to assess the significance of variables in individual models. The methods are model-agnostic so that we can assess different models similarly. Plots depict the importance of features through permutation importance~\citep{permutation_importance} and partial dependence plots~\citep{pdp}. These plots can be employed for any chosen variable, aiding in the evaluation of how the variable's value influences predictions.
\end{itemize}
These tabs have been depicted and further described in Appendix~\ref{app: components}.Figures in the following sections of the article are derived directly from the \textit{cattleia} application. Hence the font and background specific to the app's settings.




\section{Use case analysis}  \label{sec: use cases}

The \textit{cattleia} is a vital tool for data scientists to assist them in their daily work. There is a noticeable need for more tools designed to explain ensembles of models. We provide the application that requires only data and a pre-trained ensemble, providing data scientists with a comprehensive dashboard. In this Section, we present the different use case scenarios of problems one can encounter. Then, we propose a solution within \textit{cattleia} app and present real-life examples with brief analyses that we obtain in the introduced application. 

\subsection{Component models evaluation}

\textbf{Problem.} Ensembles typically comprise models of varying quality. Including models weaker in performance measures could benefit the ensemble by offering effective prediction patterns on complex data samples. While this might not always significantly improve this specific performance measure, it is crucial to remember that this largely depends on used data. 

Hence, there is a necessity for a tool that allows the examination of the performance and predictive scheme of the ensemble model and its individual components,  thereby enabling easier control over the quality of the final output.

An equally important challenge in training effective models is their generalization of unseen data. Consequently, comparing models' performance on both the training and testing sets is essential to verify if ensembles are stable and their effectiveness is truly generalized.
\vskip 0.1in
\noindent \textbf{Solution.} The performance of each component model and the ensemble as a whole can be easily scrutinized via the evaluation metrics tab. With the set of measures for classification and regression tasks, it is easy to conduct a detailed quality analysis of every model in the ensemble. 

\begin{figure}[!ht]
    \centering
    \includegraphics[width=\textwidth]{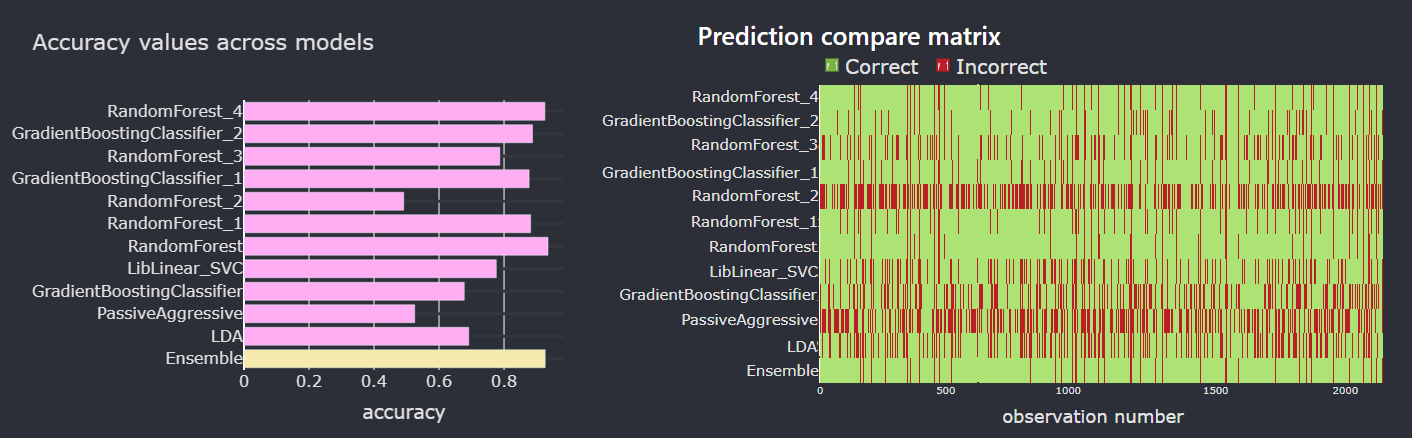}
    \caption{Bar chart showing the value of accuracy and prediction compare matrix showing the difference between predicted and true value of observation for each model.}
    \label{figure: evaluation}
\end{figure}

Each component model also has a unique prediction pattern for the provided data. This can be verified by viewing the prediction compare matrix (see Figure~\ref{figure: evaluation}). With the aid of this visualization, one can identify data areas or records that most models find challenging to predict but are correctly classified by a minority of models.

Stability and generalization among different datasets can be confirmed by supplying different datasets with models to our app. Despite the metric value typically dropping on the testing set for single models, it is noteworthy that the ensemble's performance on test data is usually more stable. This implies that the created ensemble has a more robust generalization capability, a highly sought-after trait.

\vskip 0.1in
\noindent \textbf{Example.} The example in Figure~\ref{figure: evaluation} showcases the score of an ensemble made up of 11 component models. As can be observed on the accuracy plot, each model's performance varies when comparing the evaluation metric with the ensemble having the highest value. 
On the prediction compare matrix, the misclassified areas of the set are red-colored. We can see that the quality of models is related to their correct classification probabilities. Out of all base models with many mislabeled observations, the ensemble extracted only the best patterns, resulting in an improved output.

\subsection{Diversity examination}

\textbf{Problem.} Building competitive ensembles requires including models that produce distinct predictions. This is essential because having a diverse range of models with varying methodologies makes it easier to harness the predictive power of individual models on specific data observations. The similarity analysis between component models is not trivial, as special measures are needed to define and compare their likeness.

Moreover, it is imperative to analyze the diversity and complementarity, which are highly desirable characteristics for a set of models within an ensemble. Assessing these features post-training leads to a more profound comprehension of ensemble selection and the decision-making process.  

\begin{figure}[!ht]
    \centering
    \includegraphics[width=\textwidth]{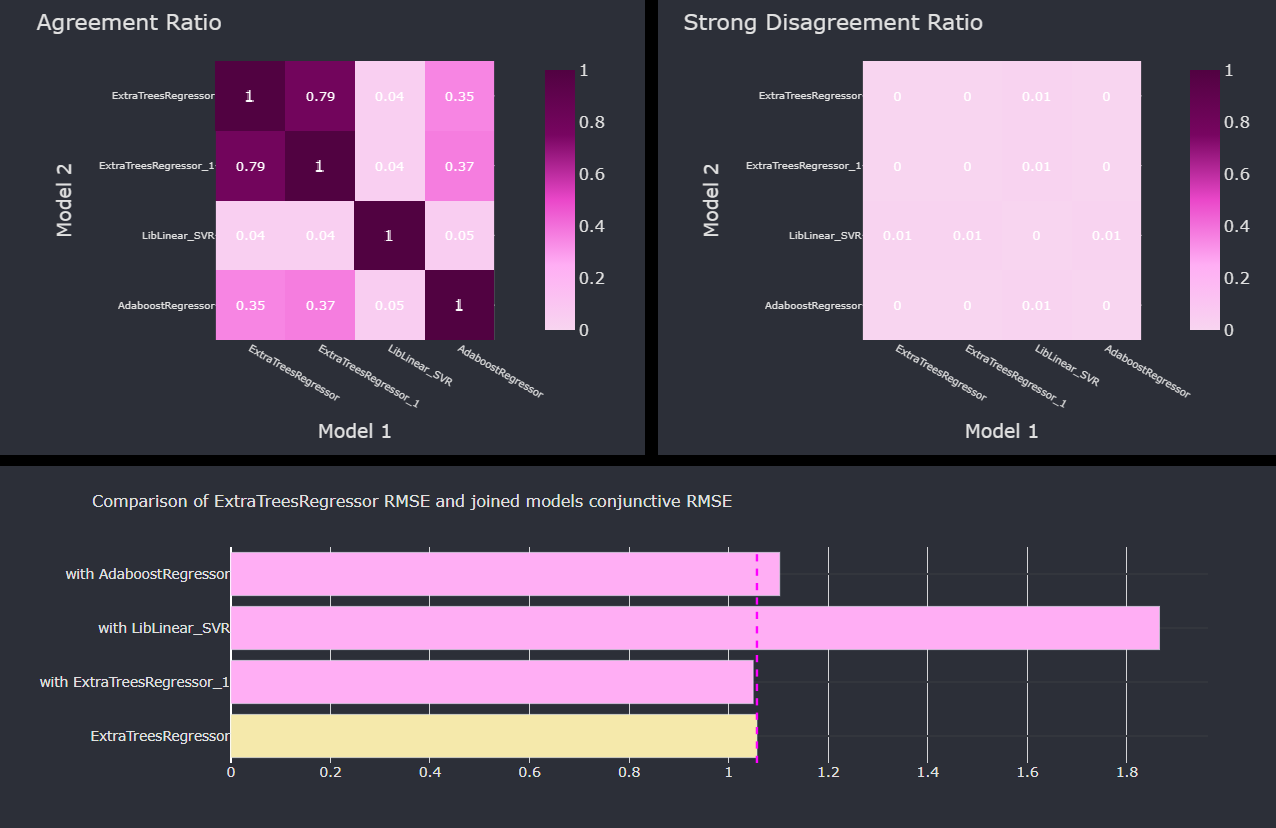}
    \caption{Strong disagreement, agreement ratio showing prediction consistency, and conjunctive accuracy showing the difference in performance of the combined two models.}
    \label{figure: complementarity}
\end{figure}
\vskip 0.1in
\noindent \textbf{Solution.} The compatimetrics tab evaluates the similarity between model predictions, providing valuable insights into model likeness. By definition, these measures capture different aspects of~diversity among model predictions, but all compatimetrics are effective in assessing the collective performance of models. This allows the identification of groups of models that synergize well or pinpoint components that hinder the final prediction. Specifically, conjunctive metrics are ideal for determining whether merging the predictions of two models enhances or degrades the overall outcome.

\vskip 0.1in
\noindent \textbf{Example.} Analyzing the agreement and strong disagreement ratio matrices presented in Figure~\ref{figure: complementarity} shows that not all model pairs are identical, with some models demonstrating a significantly low agreement ratio with others. It is worth mentioning that similarity between models is also intrinsically linked to their overall performance because better-performing models are more likely to have a higher level of similarity. 

On a bar plot presenting conjunctive RMSE, one can compare the performance of a single model against joined models. In the case of the Extra Trees Regressor model, the metric improvement by~averaging the prediction of the model pair was not achieved. We can assume that this particular model is an accurate predictor by itself. 

This analysis suggests that dissimilarity can be beneficial in specific committee scenarios. Conversely, it is common for models to exhibit a high degree of similarity; however, more is needed to guarantee effective predictive performance. 

\subsection{Sensitive data}

\textbf{Problem.} Ensuring fairness and equality is a pivotal aspect of many applications of machine learning models. Algorithms should not exhibit any form of discrimination. Therefore, before deploying a model in a production setting, it must be understood thoroughly to mitigate any potential discriminatory behavior. 
\vskip 0.1in
\noindent \textbf{Solution.} Using the XAI techniques, we can quickly evaluate the significance of a particular variable for a specific model. By observing feature importance and partial dependence plots, we can analyze how much impact a given change in a variable has on the model's predictions. 

\begin{figure}[!ht]
    \centering
    \includegraphics[width=\textwidth]{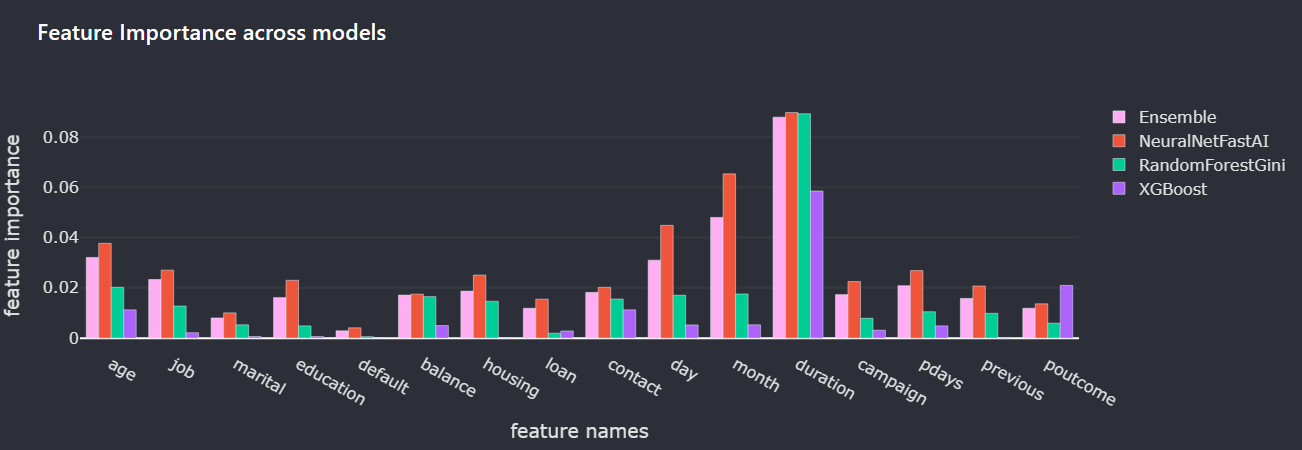}
    \caption{The feature importance plot shows the variables' influence on the model's predictions.}
    \label{figure: fairness}
\end{figure}

With this, we can detect inappropriate behavior of a particular model and correct it by~adjusting the variables it was trained on or modifying the impact of that individual model on the overall ensemble output. 
\vskip 0.1in
\noindent \textbf{Example.} The use-case presented in Figure~\ref{figure: fairness} shows that for this particular ensemble, the majority of~component models are consistent in choosing the most important variables. There is a high chance that features named job, marital, and education could be recognized as sensitive. In this case, we can observe that they do not play a crucial role, so we assume that the presented model meets the principle of fairness.

\subsection{Weights analysis and modification}

\textbf{Problem.} The process of assigning weights to model predictions is paramount in deriving the final prediction of an ensemble. These weights dictate the extent to which each component prediction will affect the ensemble's performance, making the weight distribution analysis vital for understanding ensembles' functioning.

Models with higher prediction quality are allocated greater weights. However, there can be instances where a model with sub-optimal metrics is assigned the highest weight. While this may initially seem questionable, it is important to note that an ensemble favoring a weaker model can still perform optimally by incorporating beneficial signals from the other models. While we assume that the model selection carried out by AutoML packages is optimal, there are instances where the ensemble's performance can be improved by excluding redundant component models. 
\vskip 0.1in
\noindent \textbf{Solution.} The weight modification tool allows users to analyze and experiment with the weight distribution among models within an ensemble, serving as an invaluable asset for customization and optimization. It proves particularly handy when the goal is to improve a single metric, all metrics, or optimize the ensemble. The capacity to enhance metrics largely depends on the trained model and the dataset. 

The weight modification tool serves as a solution for identifying anomalies such as unfair, poor-quality components or redundant models. We can test whether the omission of such models negatively affects the teams' performance and to what extent. This approach offers a convenient way to adjust ensembles post-training, eliminating the need to retrain models with more complex parameter settings.

\begin{figure}[!ht]
    \centering
    \includegraphics[width=\textwidth]{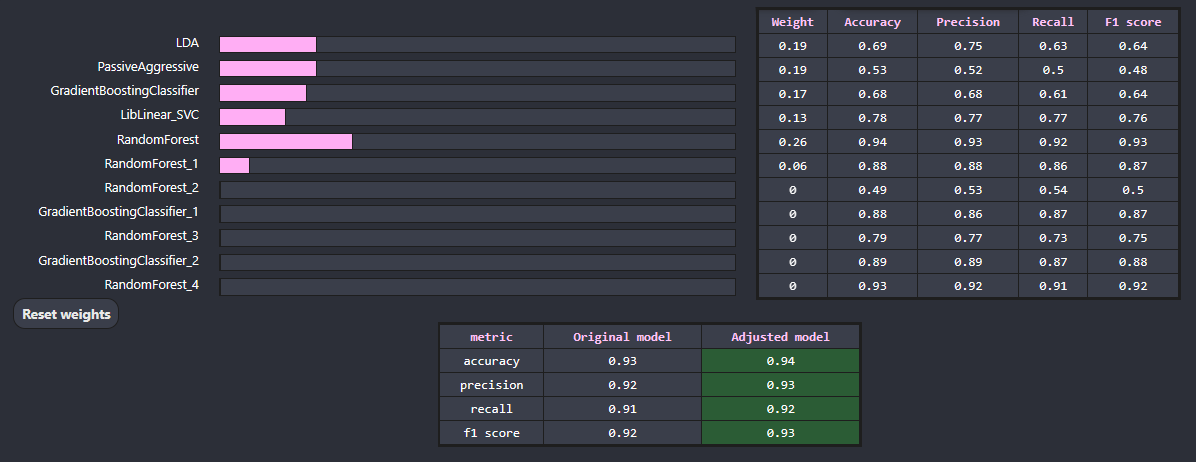}
    \caption{Weight modify tool showing difference of ensemble model performence after changing the weights of individual component models.}
    \label{figure: weights-example}
\end{figure}

\vskip 0.1in
\noindent \textbf{Example} As depicted in Figure \ref{figure: weights-example}, trimming the ensemble from eleven models to just six while improving all available metrics is possible. While such occurrences are rare, they highlight the versatility of the weight modification tool. It allows users to understand each model within the committee's significance easily. It serves as an effective means to discover a superior, smaller, and less time-consuming weighted ensemble for the chosen data. Nevertheless, it is essential to~highlight that the observed improvement is limited to the considered dataset. To avoid the risk of~overfitting, assessing how well this modified ensemble performs on the test set is essential, ensuring a reliable and broad application.

\newpage
\subsection{Summary of use cases}
Analyzing real-life use cases shows that the \textit{cattleia} could significantly enhance our understanding of ensemble models. Our application enables us to delve into the decision-making process of~constructing them, scrutinize the performance of each component model rather than just the ensemble as a whole, and investigate prediction patterns, similarities, and their impacts on the final prediction.

It discourages the use of models that are not fully understood and illuminates the reasoning behind committees. This multifaceted analysis offers a fresh perspective on analyzing models trained using AutoML packages. At the same time, it provides a clear rationale for planning in~real-world application scenarios – aspects that are always crucial when using AI. 

\section{Conclusion}



Currently, there needs to be more tools designed to help explain committee models. As the field of AutoML rapidly evolves, humans increasingly depend on algorithms that handle the entire model training pipeline. This leaves users with little control over the process, highlighting the need for a tool to take on the subsequent step - explaining AutoML models. 
For this purpose, we have created \textit{cattleia}, which could significantly enhance our understanding of the final output of most AutoML frameworks -- ensemble models.
Our app is intuitive and easy to use; therefore, it fits into the process of creating AutoML models. Its transparent visualizations make it easy for machine learning beginners and experienced data scientists to uncover valuable insights and understand the impact of various factors on model performance. This empowers users to make informed data-driven decisions. Our solution addresses the increasing demand for understanding machine learning models and the rising popularity of AutoML. 

The analysis of ensemble models in our app can be conducted from multiple perspectives, such as analysis of the performance of individual models using metrics, assess the significance of variables in individual models, investigation of the effect of modifying the weights on the performance of the model, and compare component models among themselves. The analysis can be viewed at varying levels of granularity, ranging from the entire ensemble model, pairs of models within the ensemble, down to individual models, and further refined to the examination of single variables and specific observations. This diverse approach to investigating the performance of~model committees makes it possible to answer many of the key questions when working with AutoML models.

\section{Future Works} \label{Future work}
Despite its versatile features, our application is not exempt from limitations, which encompass both functional aspects and factors that can impact the user base. The main limitation we want to work on shortly is the number of frameworks supported by the application. The \textit{cattleia} supports three of the most popular frameworks, and we plan to continue its maintenance and extend its support to future AutoML frameworks. We also aim to enhance our analysis and visualization capabilities. An interesting development direction for \textit{cattleia} is to increase the number of metrics used for model comparison and incorporate more XAI methods for a better understanding of model performance. Additionally, we plan to broaden the definitions of compatimetrics. Currently, our compatimetrics compare two models at a time, but we intend to revise them to simultaneously compare multiple models.

An important drawback of our application is that the user has to install it manually to use it. To remedy this, we plan two solutions. The first is to host the application, while the second is to create a package containing all the application components. The advantage of this solution is that the users can choose only those analyses and visualizations that interest them. 


\section{Broader Impact Statement} \label{sec: broader impact}

%
%
%
%
%


The \textit{cattleia} tool is versatile and has the potential to be used across various applications where supervised machine learning models are employed. Our main goal is to elucidate ensemble models that are developed by AutoML frameworks, thereby enhancing the comprehension of individual decisions as well as the underlying models. Using such a tool carries numerous benefits, among which includes the amplification of transparency in applications that are critical to decision-making. We are addressing the current void of unexplained ensemble models in areas such as medicine, finance, and many others. By offering an examination of ensembles and their base models, we are enriching the analysis.

We believe our application will have a substantial impact on the data science community by~aiding in the mitigation of fairness as users of the application can identify disturbing relationships in committees and models they are comprised of. This suggests that \textit{cattleia} can act as a remedy for the misuse of sensitive data, thereby enhancing the integrity of models.

Conversely, the use of such a dashboard by individuals lacking the necessary domain knowledge could be detrimental. Furthermore, a lack of understanding of the methods employed in our application, primarily compatimetrics and the weight modification tool, could result in misconceptions about ensembles. This could heighten the risk of the model or its outputs being misused. We are addressing this issue by including clear annotations that correspond to visualisations, enabling users to observe and analyse the results while simultaneously comprehending them thoroughly.

\begin{acknowledgements}

\end{acknowledgements}


\bibliography{references}

\begin{thebibliography}{}

\bibitem[Altmann et~al., 2010]{permutation_importance}
Altmann, A., Toloşi, L., Sander, O., and Lengauer, T. (2010).
\newblock {Permutation importance: a corrected feature importance measure}.
\newblock {\em Bioinformatics}, 26(10):1340--1347.

\bibitem[Breiman, 1996]{breiman96_bagging}
Breiman, L. (1996).
\newblock Bagging predictors.
\newblock {\em Machine Learning}, 24(2):123--140.

\bibitem[Caruana et~al., 2006]{caruna2}
Caruana, R., Munson, A., and Niculescu-Mizil, A. (2006).
\newblock {Getting the Most Out of Ensemble Selection}.
\newblock In {\em Proceedings of Sixth International Conference on Data Mining}, pages 828--833.

\bibitem[Caruana et~al., 2004]{caruna1}
Caruana, R., Niculescu-Mizil, A., Crew, G., and Ksikes, A. (2004).
\newblock Ensemble selection from libraries of models.
\newblock In {\em Proceedings of the Twenty-First International Conference on Machine Learning}, page~18.

\bibitem[Chatzimparmpas et~al., 2020]{viz_interpreting}
Chatzimparmpas, A., Martins, R.~M., Jusufi, I., and Kerren, A. (2020).
\newblock A survey of surveys on the use of visualization for interpreting machine learning models.
\newblock {\em Information Visualization}, 19(3):207–233.

\bibitem[Dai et~al., 2017]{DAI201775}
Dai, Q., Ye, R., and Liu, Z. (2017).
\newblock Considering diversity and accuracy simultaneously for ensemble pruning.
\newblock {\em Applied Soft Computing}, 58:75--91.

\bibitem[Drozdal et~al., 2020]{trust_in_automl}
Drozdal, J., Weisz, J., Wang, D., Dass, G., Yao, B., Zhao, C., Muller, M., Ju, L., and Su, H. (2020).
\newblock {Trust in AutoML: exploring information needs for establishing trust in automated machine learning systems}.
\newblock In {\em Proceedings of the Twenty Fifth International Conference on Intelligent User Interfaces}.

\bibitem[Erickson et~al., 2020]{agtabular}
Erickson, N., Mueller, J., Shirkov, A., Zhang, H., Larroy, P., Li, M., and Smola, A. (2020).
\newblock {AutoGluon-Tabular: Robust and Accurate AutoML for Structured Data}.
\newblock {\em arXiv preprint arXiv:2003.06505}.

\bibitem[Escalante, 2021]{Escalante2021}
Escalante, H.~J. (2021).
\newblock Automated machine {{Learning}}---{{A}} brief review at the end of the early years.
\newblock In Pillay, N. and Qu, R., editors, {\em Automated Design of Machine Learning and Search Algorithms}, pages 11--28.

\bibitem[Feurer et~al., 2022]{autosklr}
Feurer, M., Eggensperger, K., Falkner, S., Lindauer, M., and Hutter, F. (2022).
\newblock Auto-sklearn 2.0: Hands-free automl via meta-learning.
\newblock {\em Journal of Machine Learning Research}, 23(261):1--61.

\bibitem[Freund and Schapire, 1996]{Freund1996ExperimentsWA_boosting}
Freund, Y. and Schapire, R.~E. (1996).
\newblock {Experiments with a new boosting algorithm}.
\newblock In {\em Proceedings of the Thirteenth International Conference on International Conference on Machine Learning}, page 148–156.

\bibitem[Friedman, 2001]{pdp}
Friedman, J.~H. (2001).
\newblock {Greedy Function Approximation: A Gradient Boosting Machine}.
\newblock {\em The Annals of Statistics}, 29(5):1189--1232.

\bibitem[Kuncheva and Whitaker, 2003]{Kuncheva_diversity}
Kuncheva, L. and Whitaker, C. (2003).
\newblock Measures of diversity in classifier ensembles and their relationship with the ensemble accuracy.
\newblock {\em Machine Learning}, 51:181--207.

\bibitem[Ono et~al., 2020]{pipelineprofiler}
Ono, J.~P., Castelo, S., Lopez, R., Bertini, E., Freire, J., and Silva, C. (2020).
\newblock {PipelineProfiler: A Visual Analytics Tool for the Exploration of AutoML Pipelines}.
\newblock {\em IEEE Transactions on Visualization and Computer Graphics}, 27(2):390--400.

\bibitem[{Plotly Technologies Inc.}, 2015]{plotly}
{Plotly Technologies Inc.} (2015).
\newblock Collaborative data science.

\bibitem[Polikar, 2006]{Polikar_Emsemble}
Polikar, R. (2006).
\newblock Ensemble based systems in decision making.
\newblock {\em IEEE Circuits and Systems Magazine}, 6(3):21--45.

\bibitem[Ribeiro et~al., 2016]{why_trust}
Ribeiro, M.~T., Singh, S., and Guestrin, C. (2016).
\newblock {"Why Should I Trust You?": Explaining the Predictions of Any Classifier}.
\newblock In {\em Proceedings of the Twenty Second ACM SIGKDD International Conference on Knowledge Discovery and Data Mining}, page 1135–1144.

\bibitem[Sagi and Rokach, 2018]{Rokach_ensemble}
Sagi, O. and Rokach, L. (2018).
\newblock Ensemble learning: A survey.
\newblock {\em WIREs Data Mining and Knowledge Discovery}, 8(4):e1249.

\bibitem[Sass et~al., 2022]{deepcave}
Sass, R., Bergman, E., Biedenkapp, A., Hutter, F., and Lindauer, M. (2022).
\newblock {DeepCAVE: An Interactive Analysis Tool for Automated Machine Learning}.

\bibitem[Ting and Witten, 1997]{Ting_stacking}
Ting, K.~M. and Witten, I.~H. (1997).
\newblock Stacking bagged and dagged models.
\newblock In {\em Proceedings of the Fourteenth International Conference on Machine Learning}, page 367–375.

\bibitem[Wang et~al., 2021]{flaml}
Wang, C., Wu, Q., Weimer, M., and Zhu, E. (2021).
\newblock {FLAML: A Fast and Lightweight AutoML Library}.
\newblock In {\em Proceedings of the Fourth Conference on Machine Learning and Systems}.

\bibitem[Wang et~al., 2019]{atmseer}
Wang, Q., Ming, Y., Jin, Z., Shen, Q., Liu, D., Smith, M.~J., Veeramachaneni, K., and Qu, H. (2019).
\newblock {ATMSeer: Increasing Transparency and Controllability in Automated Machine Learning}.
\newblock In {\em Proceedings of the Thirty Eighth Conference on Human Factors in Computing Systems}, page 1–12.

\bibitem[Weidele et~al., 2020]{autoaiviz}
Weidele, D. K.~I., Weisz, J.~D., Oduor, E., Muller, M., Andres, J., Gray, A., and Wang, D. (2020).
\newblock {AutoAIViz: Opening the Blackbox of Automated Artificial Intelligence with Conditional Parallel Coordinates}.
\newblock In {\em Proceedings of the Twenty Fifth International Conference on Intelligent User Interfaces}, page 308–312.

\bibitem[Yuan et~al., 2020]{taxonomy}
Yuan, J., Chen, C., Yang, W., Liu, M., Xia, J., and Liu, S. (2020).
\newblock A survey of visual analytics techniques for machine learning.
\newblock {\em Computational Visual Media}, 7(1):3–36.

\bibitem[Zhou et~al., 2021]{Zhou2021EvaluatingTQ}
Zhou, J., Gandomi, A.~H., Chen, F., and Holzinger, A. (2021).
\newblock Evaluating the quality of machine learning explanations: A survey on methods and metrics.
\newblock {\em Electronics}, 10(5).

\bibitem[Z\"{o}ller et~al., 2023]{XAutoML}
Z\"{o}ller, M.-A., Titov, W., Schlegel, T., and Huber, M.~F. (2023).
\newblock {XAutoML: A Visual Analytics Tool for Understanding and Validating Automated Machine Learning}.
\newblock {\em ACM Transactions on Interactive Intelligent Systems}, 13(4).

\end{thebibliography}




\newpage 
\section*{Submission Checklist}

\begin{enumerate}
\item For all authors\dots
  \begin{enumerate}
  \item Do the main claims made in the abstract and introduction accurately reflect the paper's contributions and scope?
    \answerYes{The statements made in the abstract and introduction provide an accurate representation of the paper's contributions.}
  \item Did you describe the limitations of your work?
    \answerYes{See Section \ref{Future work}.}
  \item Did you discuss any potential negative societal impacts of your work?
    \answerYes{See Section \ref{sec: broader impact}.}
  \item Did you read the ethics review guidelines and ensure that your paper
    conforms to them? \url{https://2022.automl.cc/ethics-accessibility/}
    \answerYes{We read the ethics review guidelines and ensured that our paper conforms to them.}
  \end{enumerate}
\item If you ran experiments\dots
  \begin{enumerate}
  \item Did you use the same evaluation protocol for all methods being compared (e.g., 
    same benchmarks, data (sub)sets, available resources)? 
    \answerNA{The main contribution of this paper is an application. 
    Models presented in Section~\ref{sec: use cases} are the illustration of data scientist interaction with \textit{cattleia}. Analyzed ensembles come from already established AutoML frameworks and additional experiments with them are not crucial for this work.}
  \item Did you specify all the necessary details of your evaluation (e.g., data splits,
    pre-processing, search spaces, hyperparameter tuning)?
    \answerNA{}
  \item Did you repeat your experiments (e.g., across multiple random seeds or splits) to account for the impact of randomness in your methods or data?
    \answerNA{}
  \item Did you report the uncertainty of your results (e.g., the variance across random seeds or splits)?
    \answerNA{}
  \item Did you report the statistical significance of your results?
    \answerNA{}
  \item Did you use tabular or surrogate benchmarks for in-depth evaluations?
    \answerNA{}
  \item Did you compare performance over time and describe how you selected the maximum duration?
    \answerNA{}
  \item Did you include the total amount of compute and the type of resources
    used (e.g., type of \textsc{gpu}s, internal cluster, or cloud provider)?
    \answerNA{}
  \item Did you run ablation studies to assess the impact of different
    components of your approach?
    \answerNA{}
  \end{enumerate}
\item With respect to the code used to obtain your results\dots
  \begin{enumerate}
\item Did you include the code, data, and instructions needed to reproduce the
    main experimental results, including all requirements (e.g.,
    \texttt{requirements.txt} with explicit versions), random seeds, an instructive
    \texttt{README} with installation, and execution commands (either in the
    supplemental material or as a \textsc{url})?
    \answerYes{
    The main contribution of this article is the application. The application is available on the GitHub repository \url{https://anon-github.automl.cc/r/cattleia-DC83}, which includes requirements and an instructive README with installation.}
  \item Did you include a minimal example to replicate results on a small subset
    of the experiments or on toy data?
    \answerYes{It is available on the GitHub repository with application on a separate branch: \url{https://anon-github.automl.cc/r/cattleia-9D3A/examples}. The descriptions of AutoML frameworks used to create a model and datasets can be found in Appendix~\ref{appendix: data}.}
  \item Did you ensure sufficient code quality and documentation so that someone else 
    can execute and understand your code?
    \answerYes{The code includes installation instructions and documentation.}
  \item Did you include the raw results of running your experiments with the given
    code, data, and instructions?
    \answerYes{The main contribution of this paper is an application, so the results in a form of selected parts of dashboard are presented in Section \ref{sec: use cases}.}
  \item Did you include the code, additional data, and instructions needed to generate
    the figures and tables in your paper based on the raw results?
    \answerYes{The descriptions of AutoML frameworks used to create a model and datasets can be found in Appendix~\ref{appendix: data}.}
  \end{enumerate}
\item If you used existing assets (e.g., code, data, models)\dots
  \begin{enumerate}
  \item Did you cite the creators of used assets?
    \answerYes{Used models are developed using AutoML frameworks: auto-sklearn, AutoGluon and FLAML and datasets are described in Appendix~\ref{appendix: data}.}
  \item Did you discuss whether and how consent was obtained from people whose
    data you're using/curating if the license requires it?
    \answerYes{For use cases we employ publicly available datasets widely used in machine learning courses and exemplary applications. In Appendix \ref{appendix: data} we list all considered datasets with sources.}
  \item Did you discuss whether the data you are using/curating contains
    personally identifiable information or offensive content?
    \answerNA{Because we use publicly available datasets widely used in community, we believe that they are secure and do not contain offensive information.}
  \end{enumerate}
\item If you created/released new assets (e.g., code, data, models)\dots
  \begin{enumerate}
    \item Did you mention the license of the new assets (e.g., as part of your code submission)?
    \answerYes{The license is mentioned in the Section \ref{sec: framework}.}
    \item Did you include the new assets either in the supplemental material or as
    a \textsc{url} (to, e.g., GitHub or Hugging Face)?
    \answerYes{Application is available on Github repository. Models and datasets employed in use cases also are uploaded there in \texttt{.zip} or \texttt{.pkl} and \texttt{.csv} files respectively.}
  \end{enumerate}
\item If you used crowdsourcing or conducted research with human subjects\dots
  \begin{enumerate}
  \item Did you include the full text of instructions given to participants and
    screenshots, if applicable?
    \answerNA{}
  \item Did you describe any potential participant risks, with links to
    Institutional Review Board (\textsc{irb}) approvals, if applicable?
    \answerNA{}
  \item Did you include the estimated hourly wage paid to participants and the
    total amount spent on participant compensation?
    \answerNA{}
  \end{enumerate}
\item If you included theoretical results\dots
  \begin{enumerate}
  \item Did you state the full set of assumptions of all theoretical results?
    \answerNA{}
  \item Did you include complete proofs of all theoretical results?
    \answerNA{}{}
  \end{enumerate}
\end{enumerate}

\newpage
\appendix

\section{Compatimetrics} \label{appendix: compatimetrics}
In the definitions below, we assume that we have a dataset $D$ with predictor variables $X = (X_1$,~...,~$X_p$),  and target variable $Y = (y_1,~....,~y_n)$. Here, $Y$ represents the vector of observed values of the variable being predicted, while $\hat{Y}$ and $\tilde{Y}$ denotes the predicted values of particular models.
    \textbf{Mean Squared Difference (MSD).} 
        \emph{MSD} calculates the difference between two prediction vectors as a mean of quadratic difference between all data samples. The formula is:
        $$
        \mathrm{MSD} =\frac{1}{n}\sum^n_{i=1} (\hat{y}_i - \tilde{y}_i)^2.
        $$

    \textbf{Strong Disagreement Ratio (SDR).} 
        \emph{SDR} calculates the percentage of observations that were predicted strongly different by two particular models. SDR is dependent on a threshold that indicates how big the difference should be to count it as significant. This threshold can be set to various values, but in a base idea we want to use standard deviation of target variable denoted in formula as $SD(y)$.
        $$
        \mathrm{SDR} = \frac{\sum^{n}_{i=1}D_{i}}{n}, \, \mathrm{where} \, \ D_i = \begin{cases} 0, d_i < SD(y) \\ 1, d_i \geq SD(y) \end{cases}, d_i = |\hat{y}_i - \tilde{y}_i|.
        $$

    \textbf{Agreement Ratio (AR).} 
        \emph{AR} represents percentage of observations that were predicted very closely by two different models. Again, the result can vary when threshold is changed, but in a base form we set the limit value for difference to be considered small as standard deviation of target variable divided by $\xi$, where $\xi$ is a number greater than 1. In case of our solution, we set $\xi = 50$. $$
        \mathrm{AR} = \frac{\sum^{n}_{i=1}A_{i}}{n}, \,\ \mathrm{where} \,\ A_i = \begin{cases} 0, d_i > \frac{SD(y)}{\xi} \\ 1, d_i \leq \frac{SD(y)}{\xi} \end{cases}, \,\ d_i = |\hat{y}_i - \tilde{y}_i|.
        $$ 

    \textbf{Conjunctive Root Mean Squared Error (CRMSE).} 
        \emph{Conjunctive RMSE} is a metric that can evaluate joined performance of two models. With this metric, we firstly calculate mean prediction based on output from two models. The RMSE is measured on this averaged prediction. The formula can be simplified to:
        $$
        \mathrm{CRMSE} = \sqrt{\frac{1}{n}\sum^n_{i=1} (\frac{\hat{y}_i + \tilde{y}_i}{2} - y_i)^2}.
        $$

    \textbf{Two-models confusion matrix.} 
        \emph{Two-models confusion matrix} represents counts of actual values and predictions from two classification models. In this matrix there are 8 types of prediction results. For labelling we use method, where we check if particular value is actually negative or positive and then check if model made a mistake, which means that output is false, or model predicted correctly and output is true. This labels serve as a base for understanding classification compatimetrics. 

        \begin{table}[H]
        \caption{Two-models confusion matrix.}
        \begin{center}
        \begin{tabular}{|l|l|l|l|}
        \hline
        Prediction I         & Prediction II        & Actual value & Label                      \\ \hline
        1                    & 1                    & 1            & True True Positive (TTP)   \\ \hline
        1 (True prediction)  & 0 (False prediction) & 1            & True False Positive (TFP)  \\ \hline
        0                    & 1                    & 1            & False True Positive (TFP)  \\ \hline
        0                    & 0                    & 1            & False False Positive (FFP) \\ \hline
        1                    & 1                    & 0            & False False Negative (FFN) \\ \hline
        1 (False prediction) & 0 (True prediction)  & 0            & False True Negative (FTN)  \\ \hline
        0                    & 1                    & 0            & True False Negative (TFN)  \\ \hline
        0                    & 0                    & 0            & True True Negative (TTN)   \\ \hline
        \end{tabular}
        \label{tab:Two-models confusion matrix.}
        \end{center}
        \end{table}

    \textbf{Uniformity.} 
        \emph{Uniformity} is a measure of similarity between two models. It~counts the percentage of observations that both models predicted the same.
        $$
        \mathrm{Uniformity} = \frac{\mathrm{TTP + TTN + FFP + FFN}}{n}.
        $$

    \textbf{Incompatibility.} 
        \emph{Incompatibility} is an opposite of uniformity as it counts the percentage of observations that were predicted differently by models. 
        $$
        \mathrm{Incompatibility} = \frac{\mathrm{TFP + FTP + TFN + FTN}}{n} = 1 - \mathrm{Uniformity}.
        $$

\section{Application components} \label{app: components} 
\begin{figure}[H]
    \centering
    \includegraphics[scale=0.45]{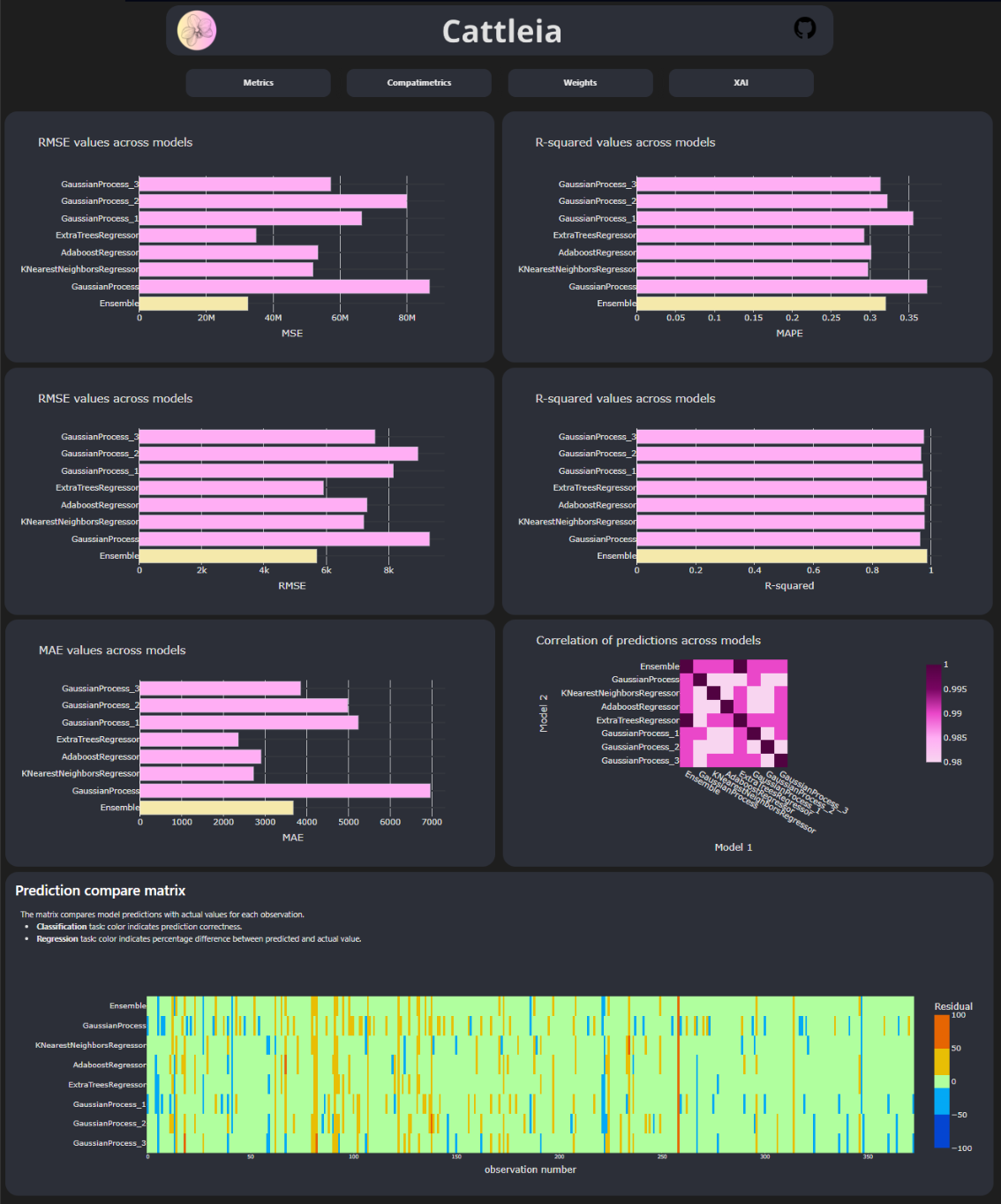}
    \caption{Example view of the metrics tab including bar charts of each metric, prediction correlation matrix, and prediction compare matrix.}
    \label{figure: metrics}
\end{figure}

The metrics tab showed in Figure~\ref{figure: metrics} contains several types of charts. The first of these is depicting evaluation measures. Bar plots show the values of each metric for the ensemble model and component model. The metrics are appropriately selected depending on the type of task. The case of regression includes metrics such as Mean Squared Error (MSE), Mean Absolute Percentage Error (MAPE), Root Mean Squared Error (RMSE), Mean Absolute Error (MAE), and $R$ Squared (\(R^2\)), while for classification: Accuracy, Precision, Recall, and F1 score. The second type of chart is a matrix showing correlations of the model's prediction. The last plot shows the differences between the predictions of each model and the true value for each observation.

\begin{figure}[H]
    \centering
    \includegraphics[scale=0.4]{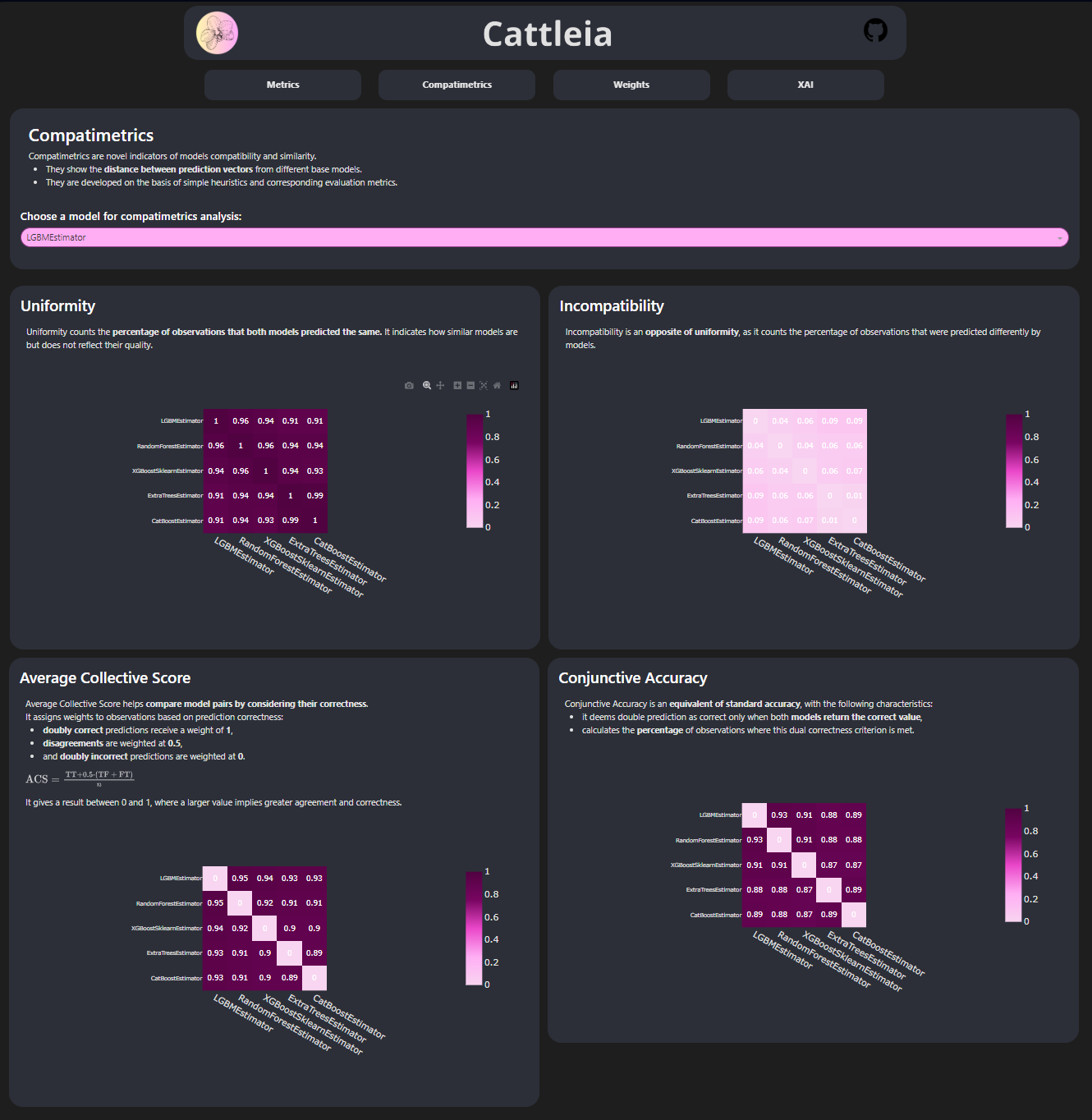}
    \caption{Part of the example compatimetrics tab view for classification includes a matrix of each compatimetrics.}
    \label{figure: compatimetrics}
\end{figure}

The compatimetrics tab shown in Figure~\ref{figure: compatimetrics} is designed to evaluate the shared performance and similarity of models that form committees.
For regressions, there are matrices of Mean Squared Difference (MSD), Root Mean Squared Difference (RMSD), Agreement (AR), and Strong Disagreement Ratio (SDR) metrics. For the model selected by the user, there are also plots of MSD and RMSD, a plot of the model's RMSE compared with the Conjunctive RMSE of the other models, and a graph showing the distribution of absolute differences in predictions between the selected model and the other models in the set.
There are matrices of Uniformity, Incompatibility, Average Collective Score, and Conjunctive Accuracy for classification. For the selected model, there are bar plots of disagreement ratio, which measures how many observations were predicted differently by two models regarding the class of the record; a stacked bar plot illustrates conjunctive metrics, a plot showing the ratio of predictions on different levels of correctness and line chart displaying the process of increasing average collective score through the whole dataset.

\begin{figure}[H]
    \centering
    \includegraphics[scale=0.45]{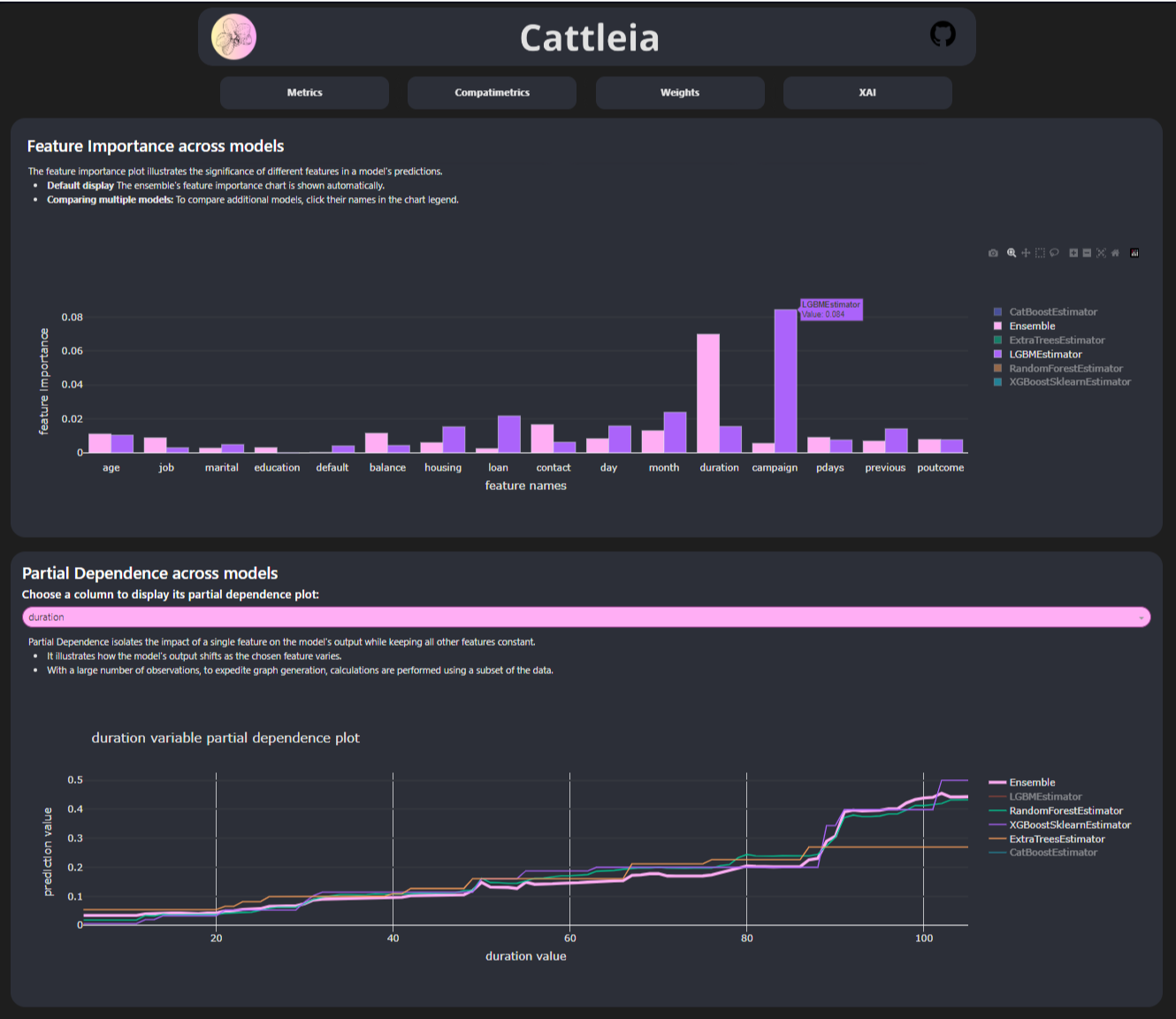}
    \caption{An example view of the XAI tab includes feature importance and partial dependence plots.}
    \label{figure: XAI}
\end{figure}

\newpage
The XAI tab shown in Figure~\ref{figure: XAI} contains two types of charts. The first is Feature importance, showing how much influence the variables have on the prediction for ensemble model and component models. The second plot is a partial dependence plot showing how, on average, a change in the value of a particular variable affects the prediction value.

\newpage

\begin{figure}[H]
    \centering
    \includegraphics[scale=0.45]{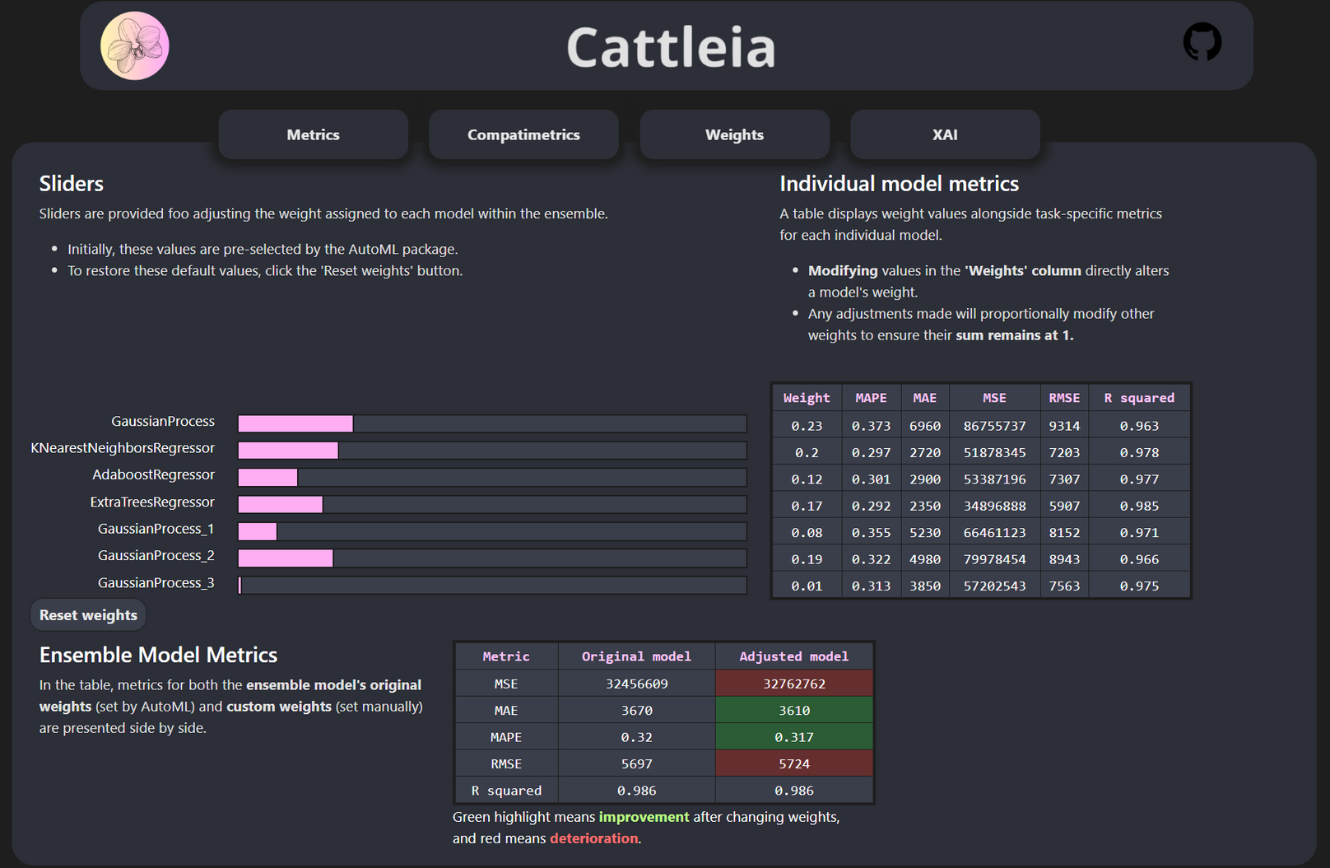}
    \caption{Example view of the weights tab, including sliders to change the weights of the models, a table showing the metrics of each model, and a table comparing model metrics after weight changes with the original model.}
    \label{figure: weights}
\end{figure}

The weights tab shown in Figure~\ref{figure: weights} allows us to see how modifying the weights of the component models of which the ensemble model is composed affects its performance. 



\section{Data} \label{appendix: data}
\begin{table}[h]
    \caption{AutoML frameworks and associated datasets used to create examples in the Section \ref{sec: use cases}.}
    \label{tab:automl_datasets}
    \centering
    \begin{tabular}{llll}
        \toprule
        \textbf{Figure} & \textbf{AutoML framework} & \textbf{Data} & \textbf{Data source} \\
        \midrule
        2 & auto-sklearn & \href{https://anon-github.automl.cc/r/cattleia-9D3A/examples/artificial_characters}{artificial\_characters} & \href{https://www.openml.org/search?type=data\&status=active\&id=1459}{OpenML - Dataset 1459} \\
        3 & auto-sklearn & \href{https://anon-github.automl.cc/r/cattleia-9D3A/examples/life_expectancy}{life\_expectancy} & \href{https://www.kaggle.com/datasets/kumarajarshi/life-expectancy-who}{Kaggle - Life Expectancy} \\
        4 & AutoGluon & \href{https://anon-github.automl.cc/r/cattleia-9D3A/examples/bank_marketing}{bank\_marketing} & \href{https://archive.ics.uci.edu/dataset/222/bank+marketing}{UCI ML Repository - Bank Marketing} \\
        5 & auto-sklearn & \href{https://anon-github.automl.cc/r/cattleia-9D3A/examples/artificial_characters}{artificial\_characters} & \href{https://www.openml.org/search?type=data\&status=active\&id=1459}{OpenML - Dataset 1459} \\
        \bottomrule
    \end{tabular}
\end{table}
\end{document}